\documentclass[%
 reprint,
 amsmath,amssymb,
 aps,
]{revtex4-2}

\usepackage{hyperref}
\usepackage{appendix}
\usepackage{algorithm}
\usepackage{algpseudocode}
\usepackage{xcolor}
\usepackage{graphicx}
\usepackage{dcolumn}
\usepackage{bm}
\usepackage{subfloat}

\begin{document}

\preprint{APS/123-QED}

\title{Scalable Sparse Regression for Model Discovery: The Fast Lane to Insight}

\author{Matthew Golden}
\email{matthew.golden@gatech.edu}
\affiliation{
 School of Physics, Georgia Institute of Technology
}

\date{\today} 

\begin{abstract}
There exist endless examples of dynamical systems with vast available data and unsatisfying mathematical descriptions. Sparse regression applied to symbolic libraries has quickly emerged as a powerful tool for learning governing equations directly from data; these learned equations balance quantitative accuracy with qualitative simplicity and human interpretability. Here, I present a general purpose, model agnostic sparse regression algorithm that extends a recently proposed exhaustive search leveraging iterative Singular Value Decompositions (SVD). This accelerated scheme, Scalable Pruning for Rapid Identification of Null vecTors (SPRINT), uses bisection with analytic bounds to quickly identify optimal rank-1 modifications to null vectors. It  is intended to maintain sensitivity to small coefficients and be of reasonable computational cost for large symbolic libraries. A calculation that would take the age of the universe with an exhaustive search but can be achieved in a day with SPRINT.

\end{abstract}

\maketitle


\section{\label{sec:level1}Introduction}
There has been a recent explosion in the application of model discovery algorithms to learn symbolic governing equations directly from temporal and spatiotemporal data. These methods rely on some form of symbolic regression \cite{bongard2007automated,schmidt2009distilling, brunton2016discovering, kaheman2020sindy} to fit data with a set of allowed symbolic operations (i.e., prescribed nonlinear functions or linear summation). Sparse regression applied to a symbolic library has been particularly successful at recovering dynamics \cite{brunton2016discovering, schaeffer2017sparse, rudy2017data, kaheman2020sindy, gurevich2021learning, alves2022data, joshi2022data, golden2023physically}. In these methods, the dynamics of a system $\dot{\bf z}$ are assumed to be a sparse linear combination of nonlinear functions $\mathcal{L} = \{ {\bf f}_n \}.$ A sparse coefficient vector ${\bf c}$ balances model simplicity with quantitative accuracy:
\begin{equation}
    \dot{\bf z} = \sum_n c_n {\bf f}_n({\bf z}). \label{eq:dynamic}
\end{equation}
Ideally, ${\bf c}$ should be chosen to be maximally descriptive while containing no irrelevant information. $\mathcal{L}$ is typically made up of polynomials or simple trigonometric functions. This is not a particularly restrictive assumption: all data is bounded and all smooth functions on bounded subsets of $\mathbb{R}^n$ are approximated arbitrarily well by polynomials \cite{stone1948generalized}. 

Model discovery has been extended to partial differential equations (PDEs) using spatiotemporal data \cite{rudy2017data, gurevich2019robust, alves2022data, joshi2022data, golden2023physically}. In these libraries, spatial gradients also enter the library. If there is a known symmetry of the system (rotational, Galilean, etc.) then $\mathcal{L}$ should contain terms that transform identically to $\dot{\bf z}$ under the action of the symmetry group \cite{gurevich2019robust, reinbold2021robust, golden2023physically}. Such symmetry considerations significantly reduce the size of the library and improve the robustness of sparse regression. 

A prevalent sparse regression paradigm for learning the dynamics is the Sparse Identification of Nonlinear Dynamics (SINDy) algorithm. In this approach, the vector is identified numerically by constructing a system of equations ${\bf G} {\bf c} = {\bf b}$ using empirical data. The $n$th column of the feature matrix ${\bf G}$ corresponds to a set of observations of ${\bf f}_n$ and ${\bf b}$ corresponds to observations of $\dot{\bf z}$. 
\begin{align}
    &{\bf G} = \begin{pmatrix}
        | & | & | \\
        {\bf f}_1 & {\bf f}_2 & {\bf f}_3 & \cdots\\
        | & | & | 
    \end{pmatrix}, 
    && {\bf b} = \begin{pmatrix} 
    | \\
    \dot{\bf z}\\
    |
    \end{pmatrix}.
\end{align}
In early model discovery efforts, ${\bf G}$ was constructed via pointwise evaluation of these library terms \cite{brunton2016discovering, rudy2017data}. Pointwise sampling is susceptible to numerical error or experimental noise, especially for terms containing derivatives. A more robust evaluation is in the weak formulation in which terms are evaluated as integrals \cite{schaeffer2017sparse, gurevich2021learning, messenger2021weak, joshi2022data, golden2023physically}. While the vector ${\bf b}$ usually corresponds to the time derivative $\dot{\bf z}$, other physically motivated terms can be used instead \cite{joshi2022data}. Once this linear system is constructed, a sparse approximate solution ${\bf c}$ is generated using sequential thresholding: the linear system ${\bf G}{\bf c} = {\bf b}$ is solved iteratively with coefficients set to zero if they are less than a threshold $|c_n| < \epsilon$, for some small $\epsilon$ \cite{brunton2016discovering, zhang2019convergence}. In practice, sequential thresholding needs surprisingly few iterations to converge.

This paradigm of sparse regression exactly minimizes a loss $\| {\bf G} {\bf c} - {\bf b} \|_2$ at each stage and then makes sparsity promoting modifications (thresholding) to arrive at a sparse coefficient vector ${\bf c}.$ This is in constrast to sparse regression methods like LASSO and elastic net, which modify the loss to contain sparsity-promoting penalties $\| {\bf G} {\bf c} - {\bf b} \|_2 + \lambda_1 \| {\bf c} \|_1 + \lambda_2 \| {\bf c} \|_2 $ \cite{santosa1986linear, zou2005regularization}. The latter approaches inevitably introduce hyper-parameters that must be tuned to obtain the desired output. Furthermore, the alignment of the $\ell_1$ sparsity-promoting penalty with the true goal of sparse regression is questionable. Why should the magnitude of coefficients matter? An $\ell_0$ penalty is certainly more desirable \cite{huang2018constructive}, although this problem is NP hard \cite{natarajan1995sparse} and tuning of a hyperparameter is still required.

Sparse regression methods face three generic difficulties: i) in principle the dynamics of a system may not be described by equations of the form \eqref{eq:dynamic} \cite{joshi2022data, golden2023physically}; ii) the symbolic expression for ${\bf b}$ is not unique if there is a degeneracy in the library;  and iii) thresholding can discard small coefficients like dissipation that are important for understanding the dynamics \cite{gurevich2021learning}. The first two problems are addressed in part by implicit-SINDy and SINDy-PI \cite{mangan2016inferring,kaheman2020sindy} where each column of ${\bf G}$ is solved for independently. 

A robust Singular Value Decomposition (SVD) based alternative was introduced during the development of the Sparse Physics-Informed Discovery of Empirical Relations (SPIDER) algorithm \cite{gurevich2019robust, gurevich2021learning, golden2023physically}. The sparse regression algorithm is an exhaustive search for optimal, rank-one modifications to a coefficient vector ${\bf c}$. This exhaustive search does not assume any right hand side ${\bf b}$, does not threshold coefficients, and can find multiple relations thereby finding any degeneracy in the library. Instead of directly solving a system of equations, it sequentially minimizes the homogeneous residual
\begin{equation}
r({\bf c}) 
\equiv 
\frac{1}{\sqrt{m}} \frac{\| {\bf G} {\bf c}\|_2}{\| {\bf c}\|_2}
\label{eq:residual}
\end{equation}
for sparse ${\bf c},$ where $m$ is the number of rows in ${\bf G}.$ All of these features can be seen directly in the example explained in Ref \cite{gurevich2021learning}: in simulated 3D turbulence the exhaustive search was able to correctly identify the pressure-Poisson equation, the energy equation, the incompressibility condition, and the Navier-Stokes equation as well as boundary conditions. The discovered equations include two spatial constraints rather than dynamical relations, and recovered three of these relations from the same scalar library displaying robustness to degeneracy. Furthermore, a viscous term with coefficient $5 \times 10^{-5}$ was identified accurately using data from two dynamically distinct regions of the flow. In another application, the exhaustive search with a modified residual found the governing equations for an experimental active nematic suspension \cite{golden2023physically}. This application displayed these same strengths: both dynamic relations and spatial constraints were identified. Six libraries were searched to obtain nine PDEs, which logically reduced to a minimal set of three governing equations.

While the exhaustive search has attractive model agnosticism and the ability to indentify accurate small coefficients, it has displayed $O(|\mathcal{L}|^4)$ scaling with library size. Calculations for $|\mathcal{L}| > 10^4$ become prohibitively expensive. Here, I propose an accelerated variant of this search algorithm:  Scalable Pruning for Rapid Identification of Null vecTors (SPRINT). This acceleration hinges on a simple correspondence: adding or removing the coefficient $c_k$ from ${\bf c}$ is equivalent to a performing a rank-1 update to the feature matrix ${\bf G} \leftarrow {\bf G} \pm {\bf g}_k \otimes {\bf e}_k$. (Here ${\bf g}_k$ is the $k$th column of ${\bf G}$ and ${\bf e}_k$ is a unit vector.) Finding the modified singular values after a rank-1 update can be done efficiently  with bisection \cite{bunch1978updating}. SPRINT comes in two flavors: SPRINT$\pm$. In SPRINT--, terms are removed from ${\bf c}$ greedily to minimize $\|{\bf G}{\bf c}\|_2$ at each iteration. In SPRINT+, one starts with a sparse guess and adds terms in the same greedy fashion. It is unnecessary to add terms until the full library is used, so in practice SPRINT+ is halted when a maximum number of terms $\|{\bf c}\|_0$ is reached. In both cases, bisection is used to quickly determine the optimal rank-1 modification to the coefficient vector ${\bf c}$. SPRINT- and SPRINT+ empirically scale like $O(|\mathcal{L}|^{3.38})$ and $O(|\mathcal{L}|^{1.65})$, respectively.

\section{The Library Catastrophe}
\begin{figure}[]
    \centering
    \includegraphics[width=0.35\textwidth]{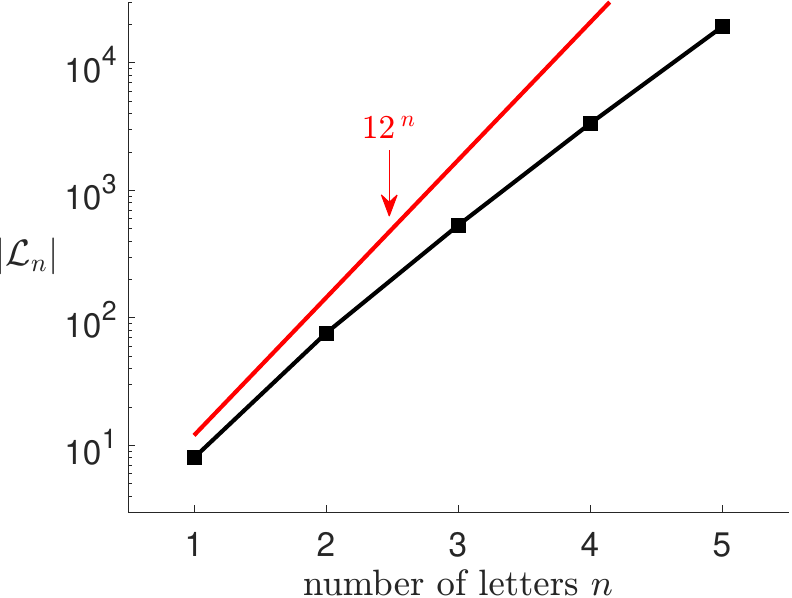}
    \caption{The size of a library $|\mathcal{L}_n|$ using 3+1D ideal MHD variables. A 12-letter symbolic alphabet is made up of eight physical fields and four partial derivatives. The library size is trivially bounded by equation \eqref{eq:Ln_upper_bound} in red. 
    Permutation symmetries make many words redundant but this does not prevent the catastrophic scaling of $|\mathcal{L}_n|$. A complete library allowing only up to 4 letters per word will require thousands of library terms. }
    \label{fig:library_scaling}
\end{figure}

Despite their success, library based methods have an unfortunate scaling with allowed complexity of library terms. This scaling is significantly reduced by symmetry considerations \cite{gurevich2021learning, golden2023physically}, but I will ignore these to illustrate the potentially catastrophic scaling of library size with complexity. Consider an ideal Magnetohydrodynamic (MHD) system with eight fields: a velocity field
$(u_x, u_y, u_z)$, a magnetic field 
$(B_x, B_y, B_z)$, a density $\rho$, and a pressure $P$. In this example, I will estimate the computational requirement for model discovery in such a system. These fields can vary in all four spacetime directions $(t,x,y,z)$, so one can take four different partial derivatives $(\partial_t , \partial_x, \partial_y, \partial_z).$ These fields and derivatives form a symbolic alphabet $\mathcal{A}$:
\begin{equation}
\mathcal{A} = \{ u_x, u_y, u_z, B_x, B_y, B_z, \rho ,P, \partial_t, \partial_x, \partial_y, \partial_z \}.
\end{equation}
Each element of this alphabet will be a \textit{letter}. A \textit{word} is a combination of letters, and its length is the total number of letters used in a word. For example, $ \rho \partial_x \rho$ is a 3-letter word. Since the number of symbols is $|A|= 12$, there can be at most $12^n$ $n$-letter words. If $\mathcal{L}_n$ is the library of words length $n$ and less, then $|\mathcal{L}_n|$ is bounded by a geometric sum of $|\mathcal{A}|^n$.
\begin{equation}
|\mathcal{L}_n| \leq  |\mathcal{A}|\frac{|\mathcal{A}|^{n} - 1}{|\mathcal{A}| - 1} \sim | \mathcal{A}|^n. 
\label{eq:Ln_upper_bound}
\end{equation}
In reality, both physicality conditions and permutation symmetries (commutativity of multiplication and partial derivatives) makes $|\mathcal{L}_n|$ smaller than this bound. The exact magnitude of the library can be found by counting words that meet the following conditions:
\begin{enumerate}    
    \item The final letter of a word cannot be a derivative: a derivative must act on something. If the final letter is a field, then this word can be evaluated.
    \item The order that fields appear in a word should coincide with the ordering $(u_x, u_y, u_z, B_x, B_y, B_z, \rho, P)$ such that $u_y u_x \wedge u_x u_y$ are redundant.
    \item Partial derivatives acting on an individual field should coincide with the canonical ordering $(\partial_t, \partial_x, \partial_y, \partial_z)$ such that $ \partial_x \partial_t u_x \wedge \partial_t \partial_x u_x$ are redundant.
    \item Products of the same field with distinct derivatives, i.e., $\partial_y \rho \partial_t \partial_x \rho$ should be ordered in a canonical way. Providing a canonical ordering is beyond the scope of this paper, but any ordering on $\mathbb{Z}^n$ is sufficient.
\end{enumerate}
The true scaling of $\mathcal{L}_n$ with $n$ for the MHD system discussed here is shown in Figure \ref{fig:library_scaling}. This $\mathcal{L}_n$ was computed exhaustively by constructing every combination of $n$ (or fewer) symbols from $\mathcal{A}$ and keeping only those that satisfy the four physicality and uniqueness conditions. By the modest size $n=5,$ one has a library with O$(10^4)$ terms. Enforcing continuous symmetry covariance, such as rotational symmetry \cite{gurevich2021learning}, can reduce the library size substantially. However, large libraries can be desirable if symmetry-breaking is suspected in the data or if no symmetry is known \textit{a priori}. Such symmetry breaking terms can appear with very small coefficients, which highlights the need for a sparse regression method without thresholding.

\section{The SPRINT Algorithm}
In the existing approach, the exhaustive search algorithm finds increasingly sparse coefficient vectors ${\bf c}$ that minimize the residual $r({\bf c})$ defined in equation \eqref{eq:residual}. Since $r({\bf c})$ is invariant under the rescaling of ${\bf c} \rightarrow \lambda {\bf c}$, we assume ${\bf c}$ is a unit vector without loss of generality. The residual is exactly minimized by performing the Singular Value Decomposition (SVD) ${\bf G} = {\bf U} {\boldsymbol \Sigma} {\bf V}^T$, where ${\boldsymbol \Sigma}$ is a diagonal matrix and ${\bf U}$ and ${\bf V}$ are rotation matrices. The diagonal elements $\sigma_i$ of ${\boldsymbol \Sigma}$ are the singular values of ${\bf G}$ and the columns of ${\bf U}$ and ${\bf V}$ are the left and right singular vectors, respectively. Efficient and stable algorithms for computing the SVD have been developed and standardized \cite{stewart1993early, anderson1999lapack}. The exact minima of $r({\bf c})$ is given by the last column of ${\bf V}$; it is the right singular vector associated with the smallest singular value $\sigma_{\textrm{min}}$ since $r({\bf c}) = \sigma_{\textrm{min}}$. This vector will be dense in general, and we are interested in sparse approximate minima. Rather than modify the residual to be sparsity promoting, we manually remove terms one at a time such that $r({\bf c})$ stays as low as possible at each iteration. Let ${\bf c}_{(k)}$ denote the coefficient vector with $k$ nonzero elements. These form approximate Pareto-optimal sets of coefficients.

This sequence of vectors are in nested sparse subspaces of the full library. Once a coefficient is set to zero, it is never allowed to become nonzero. The residuals $r_k \equiv r({\bf c}_{(k)})$ monotonically increase as $k$ decreases. One decides which term to eliminate from ${\bf c}_{(k)}$ by recomputing the SVD with each column removed ${\bf G}$. Removing the $n$th column of ${\bf G}$ before the SVD computation is equivalent to setting $c_n = 0.$ The greedy modification is selected such that $r_k$ increases the least. This means that to make a sparsification step $c_{(k)} \rightarrow c_{(k-1)}$ an SVD must be computed $k$ times. This elimination process is repeated until a single term remains. The resulting optimization curve $r_k$ shows the greedy trade-off between sparsity and quantitative accuracy in the coefficient vector ${\bf c}$. 

A selection rule for the final sparse relation is ${\bf c}_k$, where $r_{k} > \gamma r_{k+1}$ for some $\gamma>1$. The choice $\gamma=1.25$ was empirically determined \cite{gurevich2021learning}. This is the only hyperparameter of the method, but it does not change the underlying optimization curve $r_k$ or the coefficients ${\bf c}_{(k)}$. Adjusting the hyperparameter $\gamma$ has effectively no computational cost. Once a sparse relation is identified, the term with the largest contribution $\| c_n {\bf g}_n \|_2$ can be permanently removed, where ${\bf g}_n$ is the $n$th column of ${\bf G}$. This reduced library library can be searched for further relations. Searching is halted when $\sigma_{min}$ of the reduced library suggests further relations do not exist.

\begin{figure*}[t]
\includegraphics[width=0.3\textwidth]{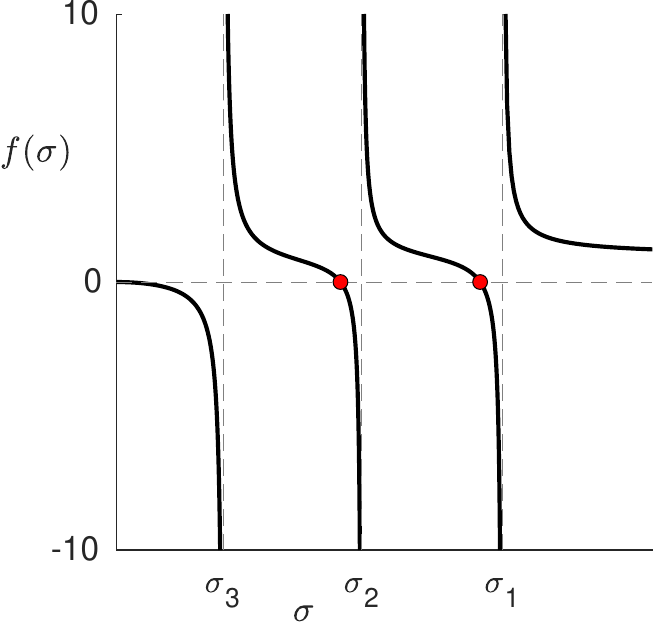}
\quad \quad
\includegraphics[width=0.3\textwidth]{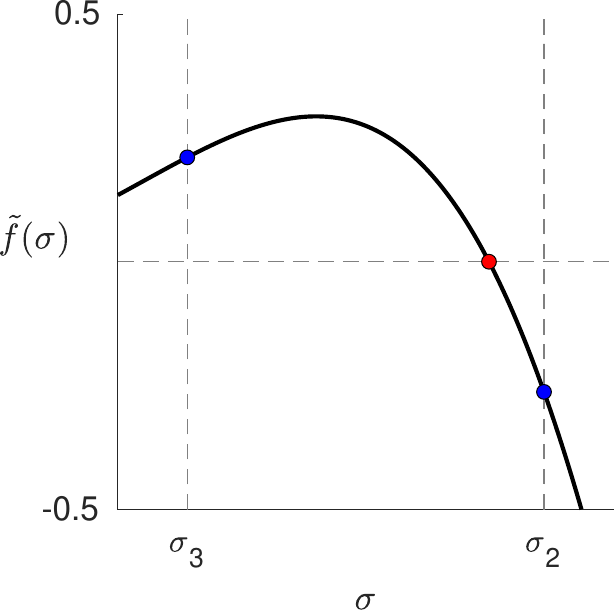}
\quad\quad
\includegraphics[width=0.3\textwidth]{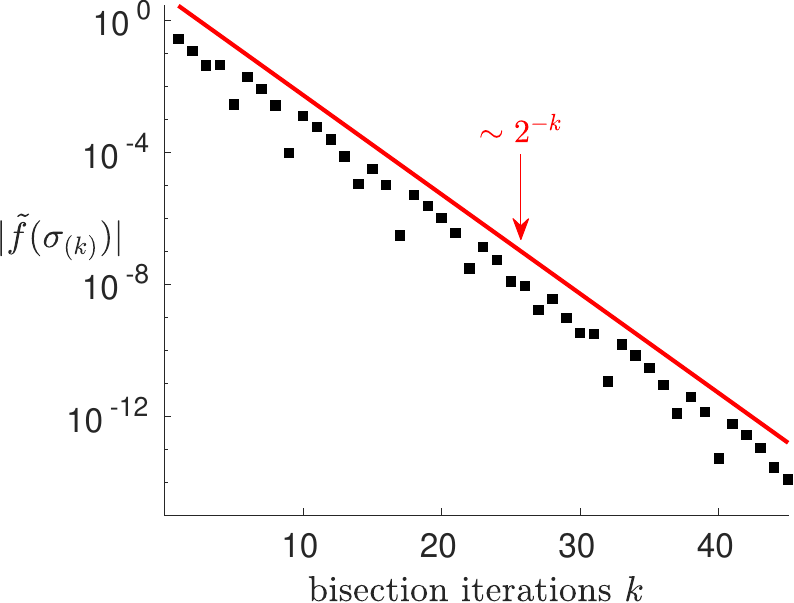}
 \caption{
An  evaluation of the secular function $f_{-}(\sigma)$ for an example matrix ${\bf G}$ in $\mathbb{R}^{3\times 3}$. The last column of the matrix is set to be removed. (a) $f_-(\sigma)$ has poles at the singular values with definitive signs: the new singular values $\sigma'_i$ will be between the previous singular values $\sigma_i > \sigma'_i > \sigma_{i+1}$ when there is no degeneracy in ${\bf G}$. The two new singular values are the roots of $f(\sigma)$ represented by red points. (b) The regularized function $\tilde{f}_{-}(\sigma)$, which is now finite at the smallest two singular values. Note that $\tilde{f}_{-}(\sigma)$ is not monotonic. The blue points represent the bounds on the singular value, and the values of $\tilde{f}_-$ at these points will always be of opposite sign. (c) The numerical convergence of the bisection. $\sigma_{(k)}$ is the $k$th iteration of bisection.
 }
 \label{fig:f_plots}
\end{figure*}

For large libraries, recomputing the SVD for every possible rank-one modification is expensive. Most computational time is spent at large $k$ when ${\bf c}_k$ is dense. Most library terms are not needed in this stage and it seems wasteful to determine the optimal way to discard thousands of unhelpful terms. There are two major considerations that speed up this method.
\begin{enumerate}
    \item All one cares about is the \textit{smallest singular value} after a rank-1 modification ${\bf G}$. Calculation of the full SVD can be avoided. 

    \item One does not need to start from scratch in the SVD evaluation. Helpful theoretical results for updating the SVD after a rank-one modification are presented in Ref \cite{bunch1978updating}. This update does require full SVD information but can be used to compute individual modified singular values quickly, which I describe below.
\end{enumerate}
Suppose that the SVD of ${\bf G}$ has already been computed ${\bf G} = {\bf U} {\boldsymbol \Sigma} {\bf V}^T$. Consider a rank-one modification ${\bf G}' = {\bf G} \pm {\bf g}_k \otimes {\bf e}_k$ such that the column ${\bf g}$ is removed or added to the matrix. The matrix determinant lemma implies that the new singular values of ${\bf G}'$ are the roots of the simple secular function $f_{\pm} (\sigma)$ \cite{bunch1978updating} with
\begin{align}  &f_{-}(\sigma) = 1 - \frac{1}{\alpha^2} \sum_{j=1}^n \frac{w_j^2}{\sigma_j^2 - \sigma^2},\\
  & f_{+}(\sigma) = 1 + \frac{1}{\alpha^2} \sum_{j=1}^n \frac{w_j}{\sigma_j^2 - \sigma^2} - \frac{1 - \|{\bf w}\|^2_2}{\alpha^2 \sigma^2},
\end{align}
where $\alpha \equiv 1/\|{\bf g}\|_2$, ${\bf w} = \alpha {\bf U}^T {\bf g}$, $n$ is the rank of ${\bf G}$ and $\sigma_j$ are the singular values of ${\bf G}.$ $f_+(\sigma)$ and $f_{-}(\sigma)$ will be used for removing and adding a column, respectively. An example of $f_{-}(\sigma)$ for a low-dimensional matrix is shown in Figure \ref{fig:f_plots}. Note the poles at the previous singular values are very useful: they imply the new singular values are uniquely placed between the old ones. The new smallest singular value $\sigma'_{\textrm{min}}\in (\sigma_{n},\sigma_{n-1})$, when removing a column, and $\sigma'_{\textrm{min}}\in (0,\sigma_{n})$, when adding a column. 

\onecolumngrid

\vspace{1cm}
\twocolumngrid

The functions $f_{\pm}(\sigma)$ have poles at these respective bounds but they are trivially removable. Indeed, regularized secular functions $\tilde{f}_{\pm}$ are better suited for numerical root-finding, i.e.,
\begin{align}
  &\tilde{f}_{-}(\sigma) = \alpha^2 \frac{(\sigma^2 - \sigma_n^2)(\sigma^2 - \sigma_{n-1}^2)}{\sigma_n^2 - \sigma_{n-1}^2} f_{-}(\sigma),\\
  &\tilde{f}_{+}(\sigma) = \alpha^2 \frac{(\sigma^2 - \sigma_n^2) \sigma^2 }{\sigma_n^2} f_{+}(\sigma).
  \label{eq:secular}
\end{align}
Let $l$ and $u$ be the lower and upper bounds of $\sigma'_{\textrm{min}}$. One can show that $f_{\pm}(l) > 0$ and $f_{\pm}(u) < 0$. By the intermediate value theorem, a root $l < \sigma < u$ exists. Bisection can be used to converge a root of these functions. One constructs a guess $\sigma_{\textrm{guess}} = (u+l)/2$ and evaluates $f_{\pm}$ at this guess. If the sign of the function is positive (negative), then replace the lower (upper) bound. Since bisection halves the candidate interval at each iteration, one can expect the error in the root at iteration $k$ to be $\delta \equiv |\sigma'_{\textrm{min}} - \sigma_{\textrm{guess}}| \propto 2^{-k}$. The value of the function will exhibit this same scaling: $f(\sigma_{\textrm{guess}}) \approx \delta f'^{\pm}(\sigma_{\textrm{min}}) \propto 2^{-k}$ if the derivative is nonvanishing at the root. This predicted scaling can be seen in Figure \ref{fig:f_plots}(c). In this example, $f_{\pm}$ will be zero to machine precision after $O(50)$ function evaluations.

SPRINT$\pm$ uses the bisection method to determine the best column modification to a matrix. After the best modification is found, the economy SVD of the modified matrix is recalculated. In SPRINT$-$, one begins with a full feature matrix and removes columns one at a time such that $\sigma_{\textrm{min}}$ increases the least at each stage. In SPRINT$+$, an initially small feature matrix has columns added such that the $\sigma_{\textrm{min}}$ decreases the most at each stage. SPRINT$+$ is not as model agnostic as SPRINT$-$ since the resulting optimization curve is heavily determined by the initial guess of sparse ${\bf c}$. One should use a dominant balance or physical domain knowledge to guess an initial relation that is refined by adding terms sequentially. While constructing this initial guess requires more thought, one does not need to add terms to the relation until the full feature matrix is recovered. One halts SPRINT$+$ once a satisfactory number of terms have been added to the relation. Indeed, most computation time in the model agnostic SPRINT$-$ is spent when the feature matrix has most of its columns, which is undesirable. SPRINT$+$ avoids this regime altogether by halting at a maximal model size.

Lastly, neither the residual $r_k$ nor the coefficient vectors ${\bf c}_k$ are affected by applying a rotation ${\bf Q}$ to the left of ${\bf G} \rightarrow {\bf Q} {\bf G}$. This rotational symmetry can be used to drastically reduce the size of ${\bf G}$ for regression. If ${\bf G}$ is a large over-determined matrix, one can compute the reduced QR factorization such that ${\bf G} = {\bf Q} {\bf R}$ \cite{trefethen1997numerical}. ${\bf R}$ is an upper-triangular square matrix and, using it in place of ${\bf G}$, eliminates any strong scaling with the number of observations in the feature matrix. This can be treated as a preprocessing step.

\section{Quantifying Algorithm Efficiency}
To quantify the efficientcy of SPRINT$\pm$, I now explore a particular example of model discovery using the Kuramoto–Sivashinsky (KS) equation, which is a classical example of a chaotic partial differential equation. The KS equation has been a canonical testing grounds for model discovery studies \cite{messenger2021weak, rudy2017data, koster2023data}. The chaotic nature of the KS equation is appealing because it provides rich timeseries data. The purpose of SPRINT is to scale to large library sizes and to learn small but important coefficients. 

In principle, such small modifications are inevitable when solving a PDE on a numerical mesh. Grid effects manifest as modifications to the governing PDE with coefficients proportional to grid spacing. To demonstrate that sparse regression can identify real small terms, I instead use a high accuracy integrator and modify the KS equation to contain several higher order nonlinearities with very small coefficients
\begin{align}
    &\partial_t u + u \partial_x u + \partial_x^2 u + \partial_x^4 u - \varepsilon \sum_{k=3}^6 \partial_x u^k = 0. \label{eq:KS_mod}
\end{align}
The small coefficient is taken to be $\varepsilon = 10^{-6}$ so that the solution closely resembles the unaltered KS trajectory. This choice is made so that the target for sparse regression is unambiguous yet it retains the flavor of identifying grid effects. 

\begin{figure*}[t]
    \includegraphics[width = 0.8\textwidth]{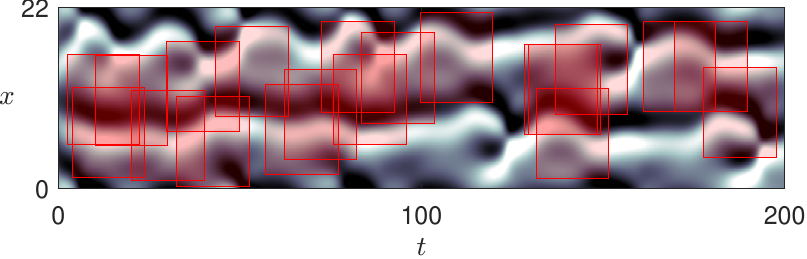}
    \caption{
    A numerical solution $u(x,t)$ to equation \eqref{eq:KS_mod}. The domain size $L=22$ is chosen to make the evolution chaotic. The red rectangles show the sizes and random distribution of weak-form subdomains used for the weak evaluation of library terms. 
    }
    \label{fig:trajectory}
\end{figure*}

I solve equation \eqref{eq:KS_mod} on a spatial domain of length $L=22$ with periodic boundary conditions for a temporal length of $T=400$. The geometry of the KS state space has been well-studied at this length and the Lyupanov exponent is $\sim 1/23$ \cite{cvitanovic2010state, edson2019lyapunov}. I use 128 gridpoints in space and 5120 in time. I approximate spatial derivatives psuedospectrally with Fast Fourier Transforms. In order to ensure that the temporal discretization error is minimal, I have used a sixth order Gauss-Legendre Runge-Kutta scheme \cite{butcher1964implicit}. This scheme is fully implicit, so the Runge-Kutta constraints are determined by iterating a Newton-Krylov scheme until their $\ell_2$ error is below $10^{-9}$ \cite{zhigunov2023exact}. Achieving this accuracy requires O(10) Newton iterations per timestep. The initial condition is $u_0(x) = \cos(3\tilde{x}) - \sin(\tilde{x})/2$ where $\tilde{x}$ is the nondimensionalized position $\tilde{x}\in[0,2\pi]$. I chose this initial condition to not have translational symmetry, although it does possess shift-reflection symmetry $u_0( \tilde{x} + \pi ) = -u_0(\tilde{x})$. The nonlinear modifications of equation \eqref{eq:KS_mod} do not respect this symmetry, so it should be violated slowly in time. The resulting dynamic field $u(x,t)$ can be seen in Figure \ref{fig:trajectory}.

\onecolumngrid

\vspace{1cm}
\twocolumngrid

I now use sparse regression to recover the modified KS equation. However, a completely general library will not be helpful for this demonstration. This is because, if equation \eqref{eq:KS_mod} is true, so are derivatives of this equation. This means sparse regression should be able to recover linearly independent relations for $\partial_t^2 u,$ $\partial_t^3 u$, $\cdots$, or some linear combination of these. Further, simple multiplication of equation \eqref{eq:KS_mod} by an arbitrary term $\phi$ produces true equations. To fix the library and only produce a single dynamical relation, I only include the time derivative once as $\partial_t u.$ I then allow for the rest of the library to contain words made from the symbolic alphabet $\mathcal{A} = \{ u, \partial_x \}$ up to length 10.
\begin{equation}
  \mathcal{L} = \{ \partial_t u \} \cup \textrm{words}_{10}( \mathcal{A} )
\end{equation}
The resulting library contains $139$ terms. Homogeneous regression on this library will either produce a dynamic closure or a spatial constraint capturing the geometry of the attractor: that is either a closure $\partial_t u = f(u, \partial_x^n u)$ or a spatial PDE $f(u, \partial_x^n u)=0$ that describes the observed data despite its dynamic nature. The numerical solution $u(x,t)$ is perturbed by additive uncorrelated Gaussian noise $\varepsilon N(\mu, \sigma)$ with amplitude $\varepsilon=10^{-7}$, mean $\mu = 0$, and a standard deviation of $\sigma = 0.218$. 

The observations of library terms are taken to be standard weak-formulation integrals over rectangular spacetime subdomains $\Omega_m$ \cite{gurevich2019robust, messenger2021weak, gurevich2021learning, golden2023physically}. Each element of ${\bf G}$ corresponds to an appropriately scaled integral over the data with a weight function $\phi$  chosen so that integration by parts is convenient. Any smooth weight function can be used, including data-driven weights \cite{golden2023physically}, but I have chosen envelope polynomials
\begin{align} 
  & G_{mn} \equiv \frac{1}{V_m S_n} \int_{\Omega_m} d\bar{x} d\bar{t}\,\, \phi f_n, \\
  & \phi(x,t) =  (1-\bar{x}^2)^8 (1 - \bar{t}^2)^8.
\end{align}
Here, $\bar{x}$ and $\bar{t}$ are rescaled coordinates that lie in the canonical interval $[-1,1]$. The nondimensionalization factors $V_m$ and $S_n$ ensure that the elements of ${\bf G}$ take on reasonable values for all library terms \cite{gurevich2021learning}. $V_m$ is the integral of the weight function.
\begin{equation}
V_m = \int_{\Omega_m} d\bar{x} d\bar{t} |\phi|.
\end{equation}
$S_n$ are combinations of characteristic scales and statistics of the data. Specifically, $S_n$ is a product of scales defined such that the scale of a undifferentiated field $u$ is its mean $\mu_u$ and the scale of a differentiated field $u$ is proportional to its standard deviation $\sigma_u$, i.e.,
\begin{align}
    &S[u] = \mu_u, && S[\partial_x^n u] = \frac{\sigma_u}{L_u^n}.
\end{align}
$L_u$ is a length scale associated with the variation of $u$. When a library term is a product of such subterms, the overall scale is the product of subscales. For reasonably smooth data, this rescaling should produce elements $G_{mn}$ that are O(1). I chose the physical size of the subdomains $\Omega_m$ to be the order of the length and time scales of characteristic process in the KS dynamics: 1024 subdomains $\Omega_m$ chosen from a uniform distribution (see Figure \ref{fig:trajectory}). 

\begin{figure*}[t]
\includegraphics[scale=0.6]{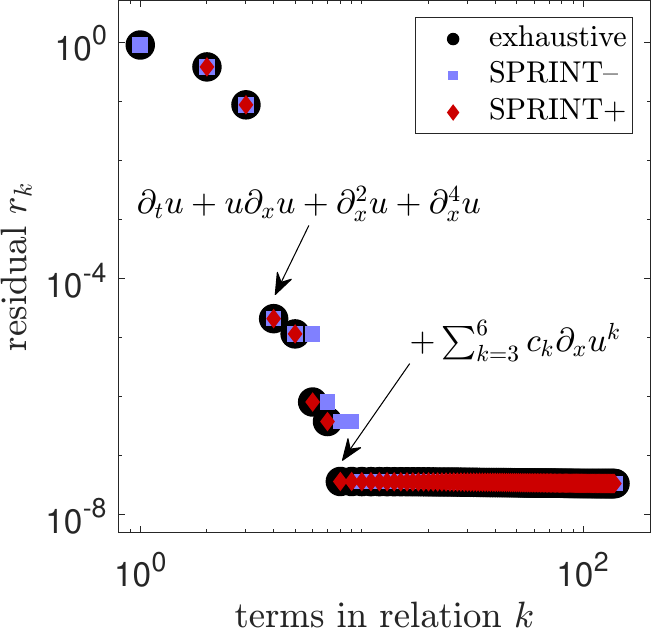} \quad
\includegraphics[scale=0.6]{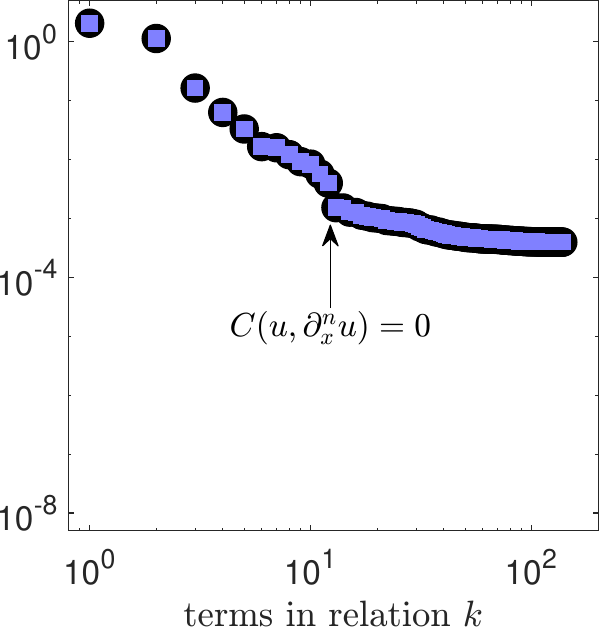} 
\caption{
Optimization curve showing the greedy residual $r_k$ as a function of nonzero terms in ${\bf c}$. The starting subspace for SPRINT+ was the $k=1$ choice of $\partial_x^2 u$ to coincide with the exhaustive search. (a) Optimization curve for the library including the time derivative $\partial_t u$. A dynamic closure is found as demonstrated by the elbow at $k=8$. (b) The optimization curve when the time derivative is \textit{not} included in the library. Instead of a dynamic closure, an approximate spatial constraint is found on the solution manifold at $k=13$.  }
\label{fig:optimization_curve}
\end{figure*}

Once I compute the matrix ${\bf G}$, I apply both an exhaustive search and SPRINT$\pm$ to produce optimization curves. The resulting curves are shown in Figure \ref{fig:optimization_curve} displaying the monotonic trade-off between quantitative accuracy and model complexity. 

\onecolumngrid
\vspace{1cm}

\twocolumngrid

\onecolumngrid

\begin{figure*}[h]
  \includegraphics[width=0.4\textwidth]{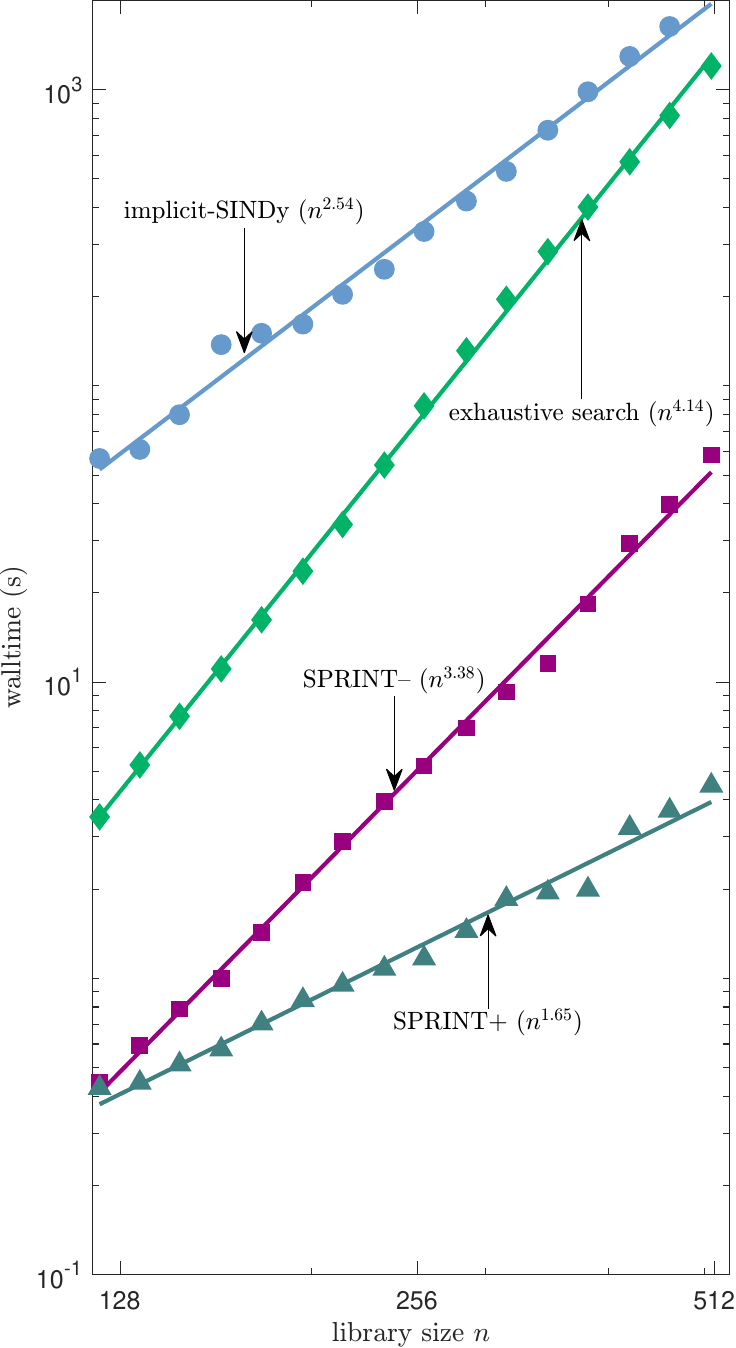}
  \quad
  \includegraphics[width=0.45\textwidth]{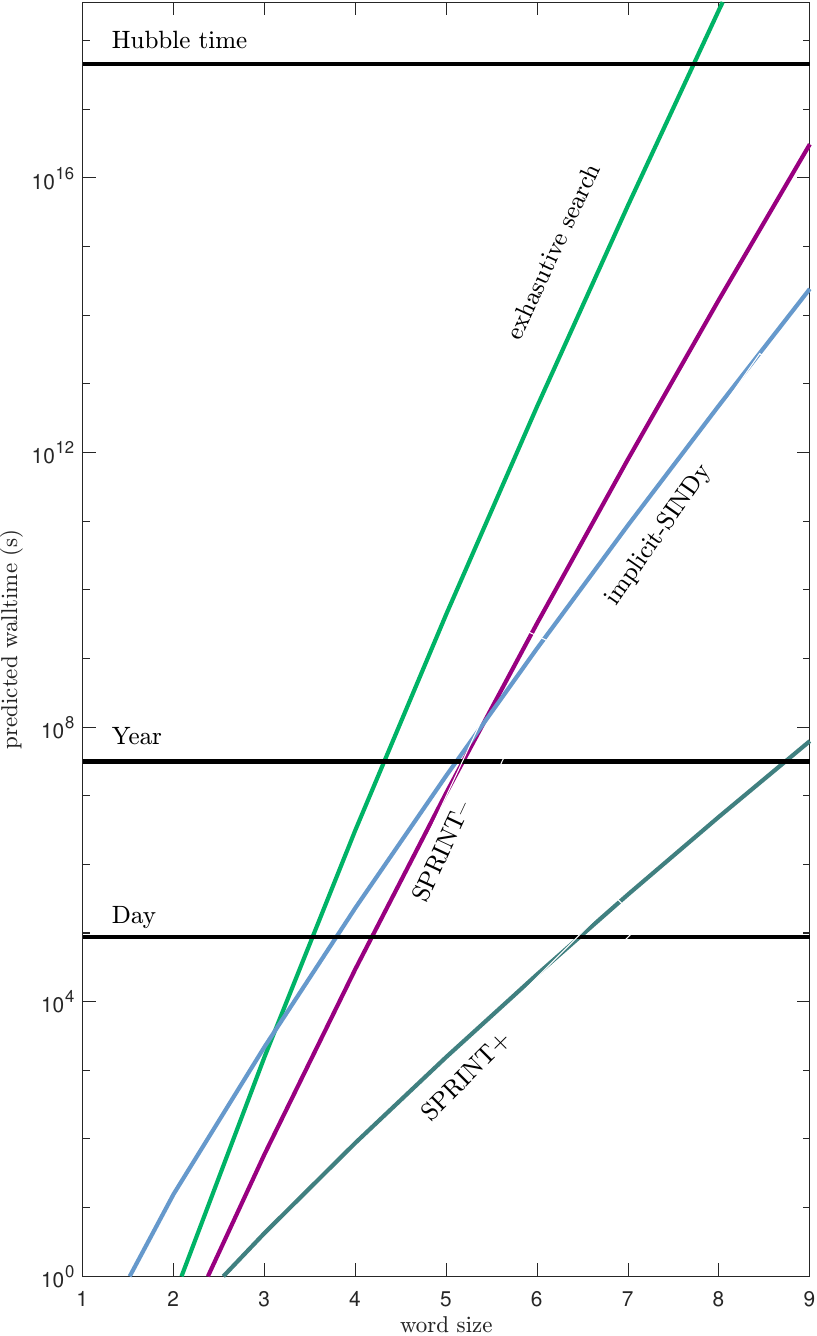}
  
  \caption{(a) Empirical walltime scaling for various sparse regression techniques. Power law exponents were empirically determined by performing a least squares fit to $\log(t) = \alpha \log(n) + \beta$ for $\alpha$ and $\beta$. Each point shows the walltime of an algorithm applied to a real square matrix with elements drawn from a uniform distribution over $[-1,1]$. The reported walltime is the mean of eight independent trials. Four algorithms are considered: the exhaustive search, SPRINT--, SPRINT+, and an implementation of implicit-SINDy available at \url{https://github.com/dynamicslab/SINDy-PI} \cite{kaheman2020sindy}.
  The exhaustive search algorithm displays the $n^{\sim4}$ scaling, while the SPRINT-- algorithm has approximately $n^{\sim3}$ scaling. These algorithms start with similar cost for small libraries, but quickly depart as the library size $n$ grows beyond O(100). (b) The extrapolated walltimes of sparse regression for the library sizes of Figure \ref{fig:library_scaling}. Calculations that would take the exhaustive search the order of a Hubble time can be accomplished in one day with SPRINT+. }
  \label{fig:scaling}
\end{figure*}
\twocolumngrid

Figure \ref{fig:optimization_curve} shows that the exhaustive search and SPRINT+ produce the same optimization curves, while SPRINT-- discards useful information too early. All three methods find an elbow in the residual at $k\approx8$. SPRINT+ and the exhaustive search exactly identify the full modified equation \eqref{eq:KS_mod} at $k=8$ as the primary elbow. A numerical instability in SPRINT-- produces an artificial elbow at k=10. All three methods recover the secondary elbow $k=4$ corresponding to the unmodified KS equation. 

The SPRINT+ method is demonstrably capable of capturing terms with small coefficients and doing so accurately. The identified coefficients of equation \eqref{eq:KS_mod} are $c_3 = 0.999\times 10^{-6}$, $c_4 = 0.993\times10^{-6}$, $c_5 = 1.001\times 10^{-6}$, and $c_6 = 1.001 \times 10^{-6}$. The ratio of residuals after including the four modifications is $r_4/r_8 = 264$. Capturing these produces a substantial change in the accuracy of the relation. 

An interesting situation arises if the time derivative is discarded from $\mathcal{L}$ so that only terms algebraic in $u$ and its spatial derivative remain. The library can then contain effective spatial descriptions of the state-space of this modified KS system. An elbow for this library can be seen in the right panel of Figure \ref{fig:optimization_curve} at $k=13$. The resulting constraint is
\begin{align}
 &C(u, \partial_x^n u) = c_1 u^2\partial_x^3 u + c_2 u\partial_x^1 u \partial_x^2 u + c_3 u\partial_x^4 u \nonumber\\
 &+ c_4 (\partial_x^1 u)^3 + c_5 \partial_x^1 u \partial_x^3 u + c_6 (\partial_x^2 u)^2 + c_7 \partial_x^5 u \nonumber\\
 &+ c_8 u\partial_x^6 u + c_9 \partial_x^1 u \partial_x^5 u + c_{10} \partial_x^2 u \partial_x^4 u + c_{11} (\partial_x^3 u)^2 \nonumber \\
 &+ c_{12} \partial_x^7 u + c_{13} \partial_x^9 u = 0.
 \label{eq:unexpected}
\end{align}
This learned constraint respects the reflection symmetry $u(x,t) \rightarrow -u(-x,t)$ of the KS equation and the initial data despite this symmetry not being enforced in the library. The modified equation \eqref{eq:KS_mod} does not possess this reflection symmetry exactly, but the effect of symmetry breaking is well below the accuracy of relation \eqref{eq:unexpected}. This highlights the versatility of homogeneous regression.

\section{Conclusion}

Bisection greatly accelerates the calculations of optimal rank-one modifications of systems of equations. These rank-one modifications include the addition and removal of coefficients in sparse regression. The two algorithms SPRINT-- and SPRINT+ presented here utilize bisection to find sparse approximate null vectors. For well conditioned matrices, SPRINT-- should exactly reproduce the exhaustive search \cite{gurevich2021learning}, although a numerical instability can lead to early removal of useful coefficients as seen in Figure \ref{fig:optimization_curve}. There can be numerical difficulties in accurately evaluating the secular equation \eqref{eq:secular} when $\sigma_{\textrm{min}}$ becomes exceptionally close to zero. If enough noise is added to the numerical solution $u$, then SPRINT-- is capable of reproducing the same optimization curve as the exhaustive search. This instability is not observed in the additive variant SPRINT+.

The only complication is that SPRINT+ needs a sparse initial guess to add terms to. This is not a problem for debugging if the suspected governing equations are known \textit{a priori}. For data with unknown governing equations, few-term dominant balances can be cheaply computed with a combinatoric search.
SPRINT+ will generally not reproduce the same curve as the exhaustive search or SPRINT-- because the optimization curve depends on the initial sparse coefficient vector. However, we see excellent performance of SPRINT+ in reconstructing the desired nonlinear dynamics with small coefficients and without enforcing the inclusion of the time derivative.

SPRINT contains no hyper-parameters other than some model selection criterion such as $r_{k-1} > \gamma r_k$ as was done in \cite{gurevich2021learning}. A different choice of such hyper-parameters does not require the costly aspects of the algorithm to be repeated. This is because the primary output is the entire optimization curve (and associated coefficients) rather than a single sparse model as in sequential thresholding or LASSO. Th ultimate choice of sparse model is up to the modeler, which lends this method to being highly interactive. There can be multiple reasonable cutoffs depending on acceptable accuracy of a relation. In Figure \ref{fig:optimization_curve} the four term KS equation provides a reasonable description of the data with a residual of $r<10^{-4}.$ 

The benefit of searching for a null vector ${\bf G}{\bf c} = 0$ rather than fitting a right hand side ${\bf G}{\bf c} = {\bf b}$ is the model agnosticism. This modification makes model discovery general purpose while not sacrificing the ability to find dynamical equations. Furthermore, the null vector approach can find single term equations like $\nabla \cdot {\bf B} = 0$ \cite{gurevich2019robust}, which the inhomogeneous method is incapable of doing. Note that implicit-SINDy \cite{mangan2016inferring} was used as a benchmark rather than SINDy-PI since SPRINT$\pm$ is investigated here as a serial algorithm. SPRINT$\pm$ can be further accelerated by parallelizing the SVD \cite{andez2008robust, jessup1994parallel} and the embarrassingly parallel bisection calculations. 

The additive variant SPRINT+ offers a staggering computational speedup compared to SPRINT-- and the exhaustive search. SPRINT+ is a promising candidate for large-library sparse regression. It displays sensitivity to small coefficients, efficient scaling with library size, and weak model agnosticism. The resulting optimization curve provides more insight than a single sparse solution. 

Since the symbolic form of modifications are unknown and potentially symmetry-breaking, it is important for sparse regression to scale well with library size. The accelerated method SPRINT presented here is an order of magnitude faster than the exhaustive search at the O(100) library. Figure \ref{fig:scaling}(a) investigates the empirical scaling of several sparse regression methods on uniform random matrices with elements in $[-1,1]$. These walltimes were measured with MATLAB's benchmarking tools. All methods display power law scaling with the library size with distinct exponents. Figure \ref{fig:scaling} displays the extropolated walltimes of various methods when considering the motivating MHD library. For small word sizes, the exhaustive search, SPRINT--, and an implementation of implicit-SINDy become prohibatively expensive. The exhaustive search in particular requires more than the Hubble time for a word size of eight, while SPRINT+ reproduces the same effective result in a day.

All code for reproducing these results is available at \url{https:/github.com/mgolden30/SPIDER}. A minimal implementation of SPRINT$\pm$ in both MATLAB and Python is available at \url{https:/github.com/mgolden30/FastSparseRegression}.

\section*{Acknowledgments}
I am thankful to Dimitrios Psaltis and Daniel Gurevich for helpful conversations.

\bibliography{bibliography}

\end{document}